\documentclass[sigconf]{aamas}  
\usepackage{balance}

\usepackage{booktabs}
\usepackage{algorithm}
\usepackage{amsmath}
\usepackage{algpseudocode}
\usepackage{graphicx}
\graphicspath{{figures}}
\usepackage{hyperref}

\usepackage{booktabs}

\setcopyright{ifaamas}  
\acmDOI{}  
\acmISBN{}  
\acmConference[AAMAS'19]{Proc.\@ of the 18th International Conference on Autonomous Agents and Multiagent Systems (AAMAS 2019)}{May 13--17, 2019}{Montreal, Canada}{N.~Agmon, M.~E.~Taylor, E.~Elkind, M.~Veloso (eds.)}  
\acmYear{2019}  
\copyrightyear{2019}  
\acmPrice{}  

\begin{document}

\title{How You Act Tells a Lot: Privacy-Leakage Attack on Deep Reinforcement Learning}  


\author{Xinlei Pan$^*$}
\affiliation{%
 \institution{UC Berkeley}
 \city{Berkeley} 
 \state{CA, USA} 
}
\email{xinleipan@berkeley.edu}

\author{Weiyao Wang$^*$}
\affiliation{%
 \institution{Duke University}
 \city{Durham} 
 \state{NC, USA} 
}
\email{ww109@duke.edu}
\author{Xiaoshuai Zhang$^*$}
\affiliation{%
 \institution{Peking University}
 \city{Beijing} 
 \state{China} 
}
\email{jet@pku.edu.cn}
\author{Bo Li}
\affiliation{%
 \institution{UIUC}
 \city{Champaign} 
 \state{IL, USA} 
}
\email{lbo@illinois.edu}
\author{Jinfeng Yi}
\affiliation{%
 \institution{JD AI Research}
 \city{Beijing} 
 \state{China} 
}
\email{yijinfeng@jd.com}
\author{Dawn Song}
\affiliation{%
 \institution{UC Berkeley}
 \city{Berkeley} 
 \state{CA, USA} 
}
\email{dawnsong@cs.berkeley.edu}
\thanks{$^*$ Indicates equal contribution.}

\begin{abstract}  
Machine learning has been widely applied to various applications, some of which involve training with privacy-sensitive data. A modest number of data breaches have been studied, including credit card information in natural language data and identities from face dataset.
However, most of these studies focus on supervised learning models.
As deep reinforcement learning (DRL) has been deployed in a number of real-world systems, such as indoor robot navigation, whether trained DRL policies can leak private information requires in-depth study.
To explore such privacy breaches in general, we mainly
propose two methods: environment dynamics search via genetic algorithm and candidate inference based on shadow policies.
We conduct extensive experiments to demonstrate such privacy vulnerabilities in DRL under various settings. We leverage the proposed algorithms to infer floor plans from some trained Grid World navigation DRL agents with LiDAR perception. The proposed algorithm can correctly infer most of the floor plans and reaches an average recovery rate of 95.83\% using policy gradient trained agents.
In addition, we are able to recover the robot configuration in continuous control environments and an autonomous driving simulator with high accuracy.
To the best of our knowledge, this is the first work to investigate privacy leakage in DRL settings and we show that DRL-based agents do potentially leak privacy-sensitive information from the trained policies.
\end{abstract}

%


\maketitle

\section{Introduction}
Recently, the machine learning field
has witnessed significant progresses on image recognition \cite{DBLP:conf/cvpr/HeZRS16},
natural language processing \cite{DBLP:journals/corr/VaswaniSPUJGKP17}, and robotic control \cite{lillicrap2015continuous}. 
However, recently machine learning algorithms have  been 
found possible to leak private information of individual 
training data \cite{DBLP:conf/sp/ShokriSSS17}. The private
information could be personal health data, transaction history
or personal photos which contain sensitive information. 
For example, in the black-box membership inference attack setting,
it is possible to determine if some individual data points
were used to train the model with only black-box access to it \cite{DBLP:conf/sp/ShokriSSS17}.
This fact indicates that personal private information 
may be leaked from the trained machine learning models.
\begin{figure}[t]
    \centering
    \includegraphics[width=0.3\linewidth]{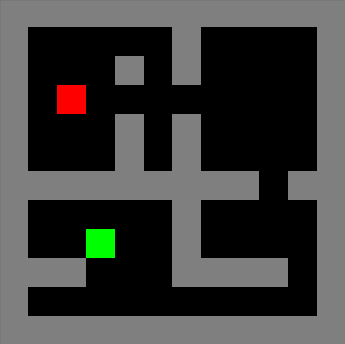}
    \includegraphics[width=0.3\linewidth]{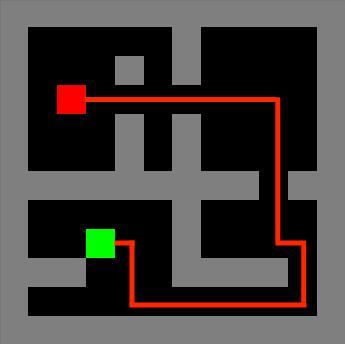}
    \includegraphics[width=0.303\linewidth]{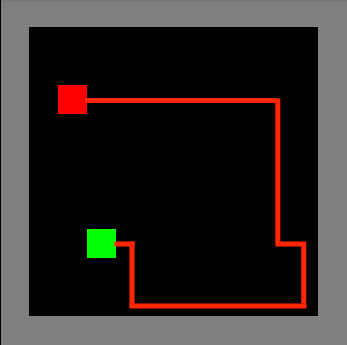}
    \caption{
    Overfitting in Grid World environment. Grid World map (left),
    agent's trained policy (middle), agent's policy when obstacles are removed (right). The agent ``memorizes'' 
     the original environment and the optimal actions even when the environment has been changed,
    indicating that the original environment transition
    dynamics can be potentially inferred by approximating the optimal policy. 
   }
    \label{fig:1}
\end{figure}

The goal of our work is to explore if similar 
issues occur in reinforcement learning (RL), especially in deep reinforcement learning (DRL). 
Deep reinforcement learning has recently achieved great successes
in solving computer games \cite{mnih2015human}, robotic control tasks
\cite{lillicrap2015continuous}, and autonomous driving 
\cite{pan2017virtual,DBLP:journals/corr/abs-1802-05313}.
Since DRL has a great potential to be applied in many real world applications which may involve personal private data in the trained model, therefore it is important to study 
the vulnerability of deep reinforcement learning to potential
privacy-stealing attacks. However, existing work on privacy in machine learning are mostly about
supervised learning models such as classification and regression models. The privacy issue in reinforcement learning algorithms has not been studied before.

We present a motivating example that demonstrates the possibility of private information leakage from a DRL policy, followed by discussions on attack strategies.
In a simple Grid World environment, the agent's task is to navigate from its current location to the goal location and avoid colliding with obstacles. In Figure~\ref{fig:1}, the gray boxes represent obstacles,
the red box denotes the agent, and the green box shows the goal of the agent. A trained reinforcement learning policy using DQN~\cite{mnih2015human} with
vision input (the aerial view of the Grid World) follows
the trajectory indicated by the red line towards the goal. Surprisingly, based on our observation, the agent follows the same trajectory even when we remove all obstacles within the frame of the Grid World. In 
this example, the trained policy will reveal the optimal actions on the original map without seeing the same map, and therefore leak the private information within the original Grid World. The explanation for this observation is that the DQN agent ``memorizes'' the training map instead of acquiring the ability to perform visual navigation. Furthermore, this information can be used to infer the map structure given that
we know the optimal action at every location. 
This motivating example shows that it is possible to infer private information from a trained deep reinforcement learning policy. 

Different from the case in supervised learning, we define the problem of stealing private information from DRL policies as inferring certain sensitive characteristics of the training environment transition dynamics given black-box access to the trained policy. We assume a powerful attacker with access to certain components of the environment including the state space, the action space, the initial state distribution and the reward function, in order to analyze the worst case scenario. Notably, such a problem of inferring transition 
dynamics is itself an ill-posed problem. Similar to inverse reinforcement learning \cite{abbeel2004apprenticeship} in which more than one reward functions can explain the observed behaviors; in our setting, it is possible that more than one transition 
dynamics can explain the given black-box policy.
As a result, in general it is not possible to infer the exact transition dynamics from the given policy. Rather, we study it as a private information leaking problem in the following two settings: first, the attacker knows nothing about the training environments but the environments can have common constraints; second, the attacker has access to a set of potential candidates of environment dynamics.


In both settings, we consider the black-box scenario which only allows us to query the policy but not the policy parameters. In the first setting, the attackers are aware of a set of explicit constraints about the transition dynamics and try to infer the transition dynamics. For instance, in the Grid World game, the attacker knows that one action can only move at most 1 step away from any location. 
In the second setting, the attacker is provided with a set of candidate robot configurations in robotic control tasks along with the trained policy in one unknown configuration, and the goal is to infer which configuration was used to perform the training. 
Such setting is motivated by the practical scenario where different robotic systems are provided in the market for further verification. Thus, by inferring the actual dynamic configuration from these accessible candidates, it is possible to tell where the robot was made or other information about the robot.
If the policy for a specific robot is hacked this way, it is possible for an attacker to infer the specific configuration of the robot and pose a great threat to the system's privacy.
In this setting, our attack trains separate \textit{shadow policies} under each of the available configurations and uses the performance on all candidate environments to train a classifier to figure out which configuration corresponds to which policy.

Our contributions are listed as follows.
\begin{itemize}
    \item To the best of our knowledge, this is the first work to conduct studies on the privacy leakage problem in DRL;
    \item We formulate the problem with goal of inferring environment dynamics under different scenarios, including when the attacker has limited knowledge about the training environments and when the attacker having access to a set of possible candidates of the training environment dynamics; 
    \item We propose two algorithms to perform the privacy attacks in DRL: approximate the optimal policy via genetic algorithm, and candidate inference via shadow policies;
    \item We perform comprehensive privacy analysis in DRL within several environments: navigation task with LiDAR perception as input, robots in continuous control environments, and autonomous driving simulator. We show that we can obtain high recovery rate in different scenarios.
\end{itemize}

\section{Related Work}


\textbf{Sensitivity of RL on Training Environments}.
It has been shown in previous work that RL policy
training is highly sensitive to training hyper-parameters, and slight variation
of these parameters can lead to significantly different results~\cite{henderson2017deep}. 
RL is also known to 
have a hard time generalizing to an unseen
domain \cite{DBLP:conf/nips/TamarLAWT16} and typically overfits
to the original training environment, though adding
randomization during training helps improve the robustness
of RL algorithms to a limited extent \cite{rajeswaran2016epopt}.
These characteristics indicate
that RL models may have implicitly memorized the training
environment and are vulnerable to attacks that try to steal
private training environment information. 

\textbf{Robust RL and Transfer Learning}. 
Previous work in robust RL
seeks to train an RL policy that can work 
in different environments with varying dynamics or visual scenes \cite{DBLP:conf/icml/PintoDSG17, pan2019risk, DBLP:conf/rss/SadeghiL17, rajeswaran2016epopt,ProceduralLevelGeneration}.
Policies trained in multiple environments have better generalization
to an unseen target domain compared with policies that
are only trained in one environment \cite{rajeswaran2016epopt}. 
Intuitively, RL models that can generalize well to different
environments may also tend to have better robustness against privacy leaking attacks. 

\textbf{Privacy-Stealing Attack Against Machine Learning Models}.
Membership inference attacks on machine learning models have been previously studied by \cite{DBLP:conf/sp/ShokriSSS17}
and further studied by \cite{song2017machine,carlini2018secret}. The attack is performed on the trained models to tell whether or not a specific data point is in the training set. On the defense side,  differential privacy for machine learning algorithms protecting training data has also been been framed and studied by \cite{PPDL, DLDP}.
However, most of the works on both sides focus on classification models, where the data are collected offline for training. In the reinforcement learning setting, the data are collected online during training, and inference about whether a single data point is used to train the model is not applicable. Instead, in this paper, we propose to infer the environment transition dynamics. One related work is privacy preserving reinforcement
learning by \cite{DBLP:conf/icml/SakumaKW08}. However, their work
is in a multi-agent reinforcement learning setting, and the private information refers to privacy between individual agent's knowledge instead of the actual training data privacy discussed in this work. 

\textbf{Inverse Reinforcement Learning}. Inverse reinforcement learning
is about inferring the reward function given expert demonstrations
\cite{abbeel2004apprenticeship,ziebart2008maximum}. In our case, we 
are given access to a black-box well-trained reinforcement learning 
policy, and the task is to infer the most likely dynamics coefficients
or the environment transition dynamics. Our work is also related with the
work by \cite{herman2016inverse}. However, this work assumes that experience data in the original environment are given, while in our case we do 
not have this assumption.

\section{Privacy-Leakage Attack on Deep Reinforcement Learning}

In this section, we first formulate our problem under the reinforcement learning setting and then propose algorithms to infer private information about the
RL training environment.

\subsection{Problem Definition}
We follow the formulation of the standard reinforcement learning problem. The environment of reinforcement learning is modeled as a Markov Decision Process (MDP), which consists
of the state space $\mathcal{S}$, the action 
space $\mathcal{A}$, the transition dynamics $\mathcal{T}$, and the reward
function $\mathcal{R}$. Usually a discount factor $\gamma$ will be applied to the 
accumulated reward to discount future rewards. The transition dynamics is defined as a probability mapping from state-action pairs to states $\mathcal{T}: (\mathcal{S}
\times\mathcal{A})\times\mathcal{S}\rightarrow[0,1]$. The goal of a reinforcement learning algorithm is
to learn a policy $\pi$ that maps state-action pairs to a probability distribution$\pi: \mathcal{S}\times\mathcal{A}\rightarrow [0,1]$, so as to maximize the expected return $\mathbb{E}_{\pi}[\sum_{t=0}^{H-1}\gamma^tr_t]$,
where $H$ is the length of the horizon.

In this work, we are interested in the problem of inferring the transition dynamics $\mathcal{T}$ of an MDP given a well trained policy $\pi$ and other components of that MDP.
Recovering the transition probability $\mathcal{T}$ is a fundamentally ill-posed problem. There could be multiple $\mathcal{T}s$ that can explain the same observed policy. Therefore, we further assume the attacker has access to prior knowledge of some structural constraints about the environment. Specifically, we consider the following two scenarios.
In the first scenario, we assume the attacker has knowledge of some explicit constraints on $\mathcal{T}$. The attacker tries to find a $\mathcal{T}$ that both satisfies the constraints and best explains the observed policy. 
In the second scenario, we assume the attacker knows a set of candidate $\mathcal{T}$s. The goal of the attacker is to determine which one of the candidates is the original transition dynamics used for training.

\subsection{Methodology}

In this section, we introduce two methods that respectively solve the two problems defined above. 

\subsubsection{Transition Dynamics Search by Genetic Algorithm} 

For the first scenario, we formulate the problem as searching for a $\mathcal{T}$ that both satisfies certain constraints and best explains the observed policy. We propose to solve such a problem with a Genetic Algorithm (GA). GA is a kind of search algorithm inspired by the process of natural selection \cite{popov2005genetic,such2017deep}. Different GAs are proposed based on different bio-inspired mutation and selection operators \cite{deb2002fast,rey1994niched}. 
In this paper, we use the basic GA to demonstrate the possibility of attack.
However, our goal is not to show
that GA is the only and the best 
method to solve the problem.

In each iteration, we maintain a population 
of transition dynamics $\{\mathcal{T}\}$ that satisfy the known constraints. 
We have the following fitness score (Equation~\ref{eq:eq2}) to characterize how similar the induced optimal policy $\pi^*_{\mathcal{T}}$ by any $\mathcal{T}$ is to the given policy $\pi_{\textrm{target}}$.  
\begin{equation}
    Score(\pi^*_{\mathcal{T}}; \pi_{\textrm{target}}) = \sum_{s \in \mathcal{S}_E}{\Delta(\pi_\textrm{target}(s), \pi^{*}_{\mathcal{T}}(s))},
    \label{eq:eq2}
\end{equation}
where $\mathcal{S}_E$ is the state space of the environment $E$, and $\pi^*_{\mathcal{T}}$ is the optimal policy under environment $E$ with transition dynamics $\mathcal{T}$.
Here $\Delta$ is a similarity metric on the action space and $\pi^{*}_{\mathcal{T}}$ refers to the optimal policy we independently obtained by training with candidate transition dynamics $\mathcal{T}$. The goal is to find a $\mathcal{T}$ such
that it maximizes this similarity score. 
Top $n$ candidates (elite population) sorted by the fitness (similarity) score are kept to the next generation. Other candidates are generated by two candidates (called parents) of the last generation. Our GA variant selects parents by randomly selecting two candidates and choosing the one with the higher score. Then a two-point crossover of the selected parents is used to generate their child candidates. Also, random mutation is applied to the child candidates. A detailed description of our GA is in Algorithm~\ref{ga}. By running such a GA algorithm, we aim to find a $\mathcal{T}_{\textrm{out}}$ that is close to the original transition dynamics that induces the provided policy $\pi_{\textrm{target}}$.

\begin{algorithm}

\caption{Genetic Algorithm for Transition Dynamics Recovery}
\label{ga}
\begin{algorithmic}
\State \textbf{Input:} 
\State \hspace*{\algorithmicindent} $\pi_{\textrm{target}}$: The provided policy 
 \State \hspace*{\algorithmicindent} $C$: The constraint for transition dynamics
\State \hspace*{\algorithmicindent} $Score$: The fitness score function 
\State \hspace*{\algorithmicindent} $L$: Size of the population 
\State \hspace*{\algorithmicindent} $n$: Size of the elite population 
 \State \hspace*{\algorithmicindent} $G$: Generation limit 
 \State \textbf{Output:} 
 \State \hspace*{\algorithmicindent} $\mathcal{T}_{\textrm{out}}$: Best transition dynamics solution found
\State $\{\mathcal{T}\}_0$ := randomly generated $L$ candidates satisfying $C$
\For{i \textbf{in} 0 \textbf{to} $G-1$}
    \State {$\{\mathcal{T}\}_{i} :=$ sort $\{\mathcal{T}\}_{i}$ according to the $Score(\pi_{\mathcal{T}}^*;\pi_{\textrm{target}})$}
    \State {$\{\mathcal{T}\}_{i+1}$ := first $n$ candidates in $\{\mathcal{T}\}_{i}$}
    \While{$\mathcal{T}_{i+1}$ is not filled up}
        \State{$\mathcal{T}_1$, $\mathcal{T}_2$ := randomly select 2 candidates in $\{\mathcal{T}\}_{i}$} 
        \State{$parent_1$ := max($\mathcal{T}_1$, $\mathcal{T}_2$) according to their score}
        \State{$\mathcal{T}_1$, $\mathcal{T}_2$ := randomly select 2 candidates in $\{\mathcal{T}\}_{i}$} 
        \State{$\textrm{parent}_2$ := max($\mathcal{T}_1$, $\mathcal{T}_2$) according to their score}
        \State{$\mathcal{T}_{\textrm{new}}$ := Crossover($\textrm{parent}_1$, $\textrm{parent}_2$)}
        \State{$\mathcal{T}_{\textrm{new}}$ := Mutation($\mathcal{T}_{\textrm{new}}$)}
        \If{$\mathcal{T}_{\textrm{new}}$ satisfies $C$}
            \State{ $\mathcal{T}_{i+1}$ = $\mathcal{T}_{i+1}\cup(\mathcal{T}_{\textrm{new}})$}
        \EndIf
    \EndWhile
\EndFor
\State{\textbf{return} $\mathcal{T}_{\textrm{out}}$: the candidate with the highest score in $\{\mathcal{T}\}_{\textrm{G}}$}
\end{algorithmic}
\end{algorithm}

\subsubsection{Candidate Inference with Shadow Policies} 

For the second scenario, we present an algorithm to perform candidate environment dynamics inference. Since the policy tends to behave differently in environments with perturbed transition dynamics, we use the given policy's performances under all candidate environments to infer which candidate it was trained on.
To build a classification model, we first construct the training dataset by training $N\times m$ shadow policies under $N$ environment dynamics, each of which is initialized with $m$ random seeds.
The policy $\pi_i^j$ is the $j$th policy trained on the corresponding candidate transition dynamics $\mathcal{T}_i$, where $i\in\{1,2,\cdots,N\},j\in\{1,2,\cdots,m\}$. Then for each of the $N\times m$ policies, we collect their episodic rewards under $N$ different environment dynamics with $k$ trials for each. 
We construct the feature vectors for each policy using the mean and variance of the episodic rewards over $k$ trials in every environment dynamics. 
Thus, the feature vector for each policy will be a vector in $\mathbb{R}^{2N}$. 
We then fit a classifier based on the feature vectors with labels corresponding to each environment dynamics candidate. During testing time, given a 
target policy $\pi_\textrm{target}$, we build the feature vector in a similar way: calculate the episodic reward
in each of the $N$ dynamics with $k$ different trials, and therefore predict which environment dynamics it was originally 
trained on.
The detailed algorithm is described in Algorithm~\ref{euclid}. 
\begin{algorithm}
\caption{Candidate Inference with Shadow Policies}\label{euclid}
\begin{algorithmic}
\State \textbf{Prepare Training Data}
\State Learn $N\times m$ shadow policies $\{\pi\}_N^m$ under $\{\mathcal{T}\}_N$; 
$\mathcal{D}=\{\}$
\For{$\pi_i^j\in\{\pi\}_N^m$}{\\
    \hspace*{\algorithmicindent} $\mathcal{D}[\pi_i^j]=\emptyset$
    \For{ $T\in\{\mathcal{T}\}_N$}{\\
        \hspace*{\algorithmicindent} \hspace*{\algorithmicindent} $R(\pi_i^j)$: obtain $k$ trials' episodic reward of policy $\pi_i^j$ in $T$\\
        \hspace*{\algorithmicindent} \hspace*{\algorithmicindent} $\mathcal{D}[\pi_i^j]$ = $\mathcal{D}[\pi_i^j]\cup$ mean and variance of $R(\pi_i^j)$
    }
    \EndFor
}
\EndFor
\State Got data pairs: rewards and dynamics label $\{(\mathcal{D}[\pi_i^j],i)\}_{i=1:N}^{j=1:m}$.
\State \textbf{Learn Classifier}
\State \hspace*{\algorithmicindent} Learn a classifier $c:\mathcal{D}[\pi_i^j]\rightarrow i.$
\State \textbf{Test}: Apply classifier $c$ on target policy $\pi_{\textrm{target}}$.
\end{algorithmic}
\end{algorithm}

\subsection{Relation to Inverse RL}
Our defined problem is related to inverse RL. For an MDP tuple ($\mathcal{S}$, $\mathcal{A}$, $\mathcal{T}$,$\mathcal{R}$, $\mathcal{\gamma}$), normal RL assumes that all 4 items are available and learns an optimal policy $\mathcal{\pi}$ to maximize the expected return.
Inverse RL (IRL) assumes that only ($\mathcal{S}$, $\mathcal{A}$, $\mathcal{T}$, $\mathcal{\gamma}$) are available. It infers $\mathcal{R}$ given optimal policy $\mathcal{\pi}$ from expert demonstrations. RL seeks to learn the optimal policy given a reward function $\mathcal{R}$, while IRL learns the reward function $\mathcal{R}$
that best supports the given policy $\mathcal{\pi}$.
Different from IRL and RL, our proposed transition dynamics recovery problem assumes that only ($\mathcal{S}$, $\mathcal{A}$, $\mathcal{R}$, $\mathcal{\gamma}$) are available. It infers $\mathcal{T}$ given some optimal policy $\mathcal{\pi}$. Our work studies how to infer the transition dynamics $\mathcal{T}$
that best supports the given policy $\mathcal{\pi}$.

\section{What Map Did You Walk On?}

In this experiment section, we present an example showing how an attacker could inversely acquire private information (the floor plan in a Grid World environment) with GA algorithm under the first scenario we defined previously.
\begin{figure}[!tbp]
  \centering
  \begin{minipage}[b]{0.4\textwidth}
    \includegraphics[width=\textwidth]{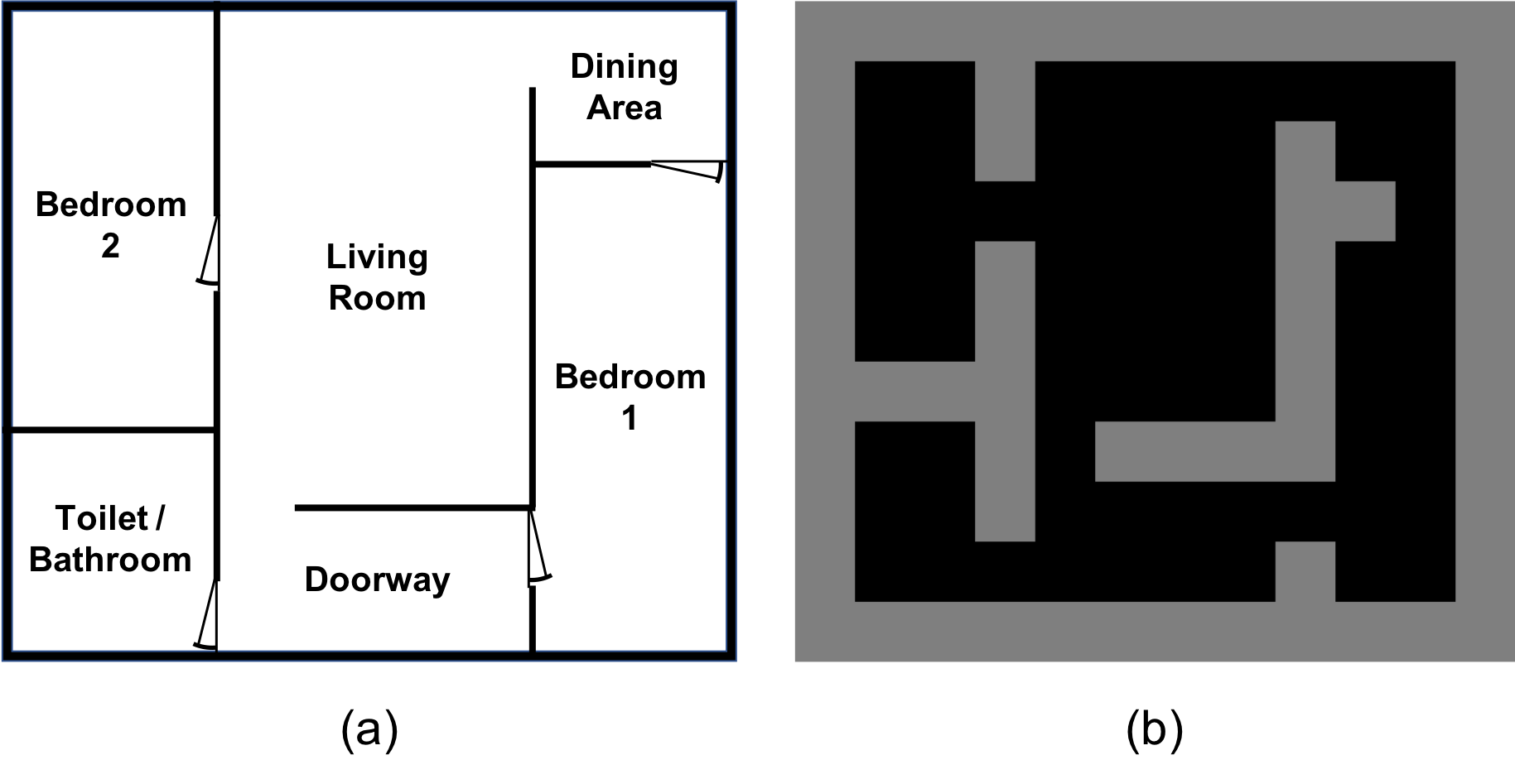}
  \end{minipage}
  \caption{An example of our abstracted grid map. (a) The original real floor plan. (b) The abstracted grid map.}
  \label{gridmapdemo}
\end{figure}

\subsection{Experimental Settings}
Performing navigation on a certain map is a fundamental problem in robotics. Recent studies try to address this problem in an end-to-end manner by using Deep Reinforcement Learning (DRL) based algorithms \cite{mirowski2016learning,zhu2017target}. As the Grid World is a very commonly used environment for testing RL algorithms, and a reasonable simplification of the navigation task, the experiments in this section are based on the Grid World environments. 

We are interested in the problem that given a well-trained DRL agent on some specific grid map, can we recover the map (or at least part of the map)? This kind of attack can pose a real threat to the practical use of RL agents, if the attacker can easily infer the floor plan structures of privacy-sensitive areas by just having access to the navigation robots' learned policy. 
We introduce the following setup to approximate the case of a navigation robot in the real world. More specifically, the grid maps are designed similarly to real floor plans. An example is presented in Figure~\ref{gridmapdemo}. 

\begin{figure*}[!tbp]
  \centering
  \begin{minipage}[b]{0.97\textwidth}
    \includegraphics[width=\textwidth]{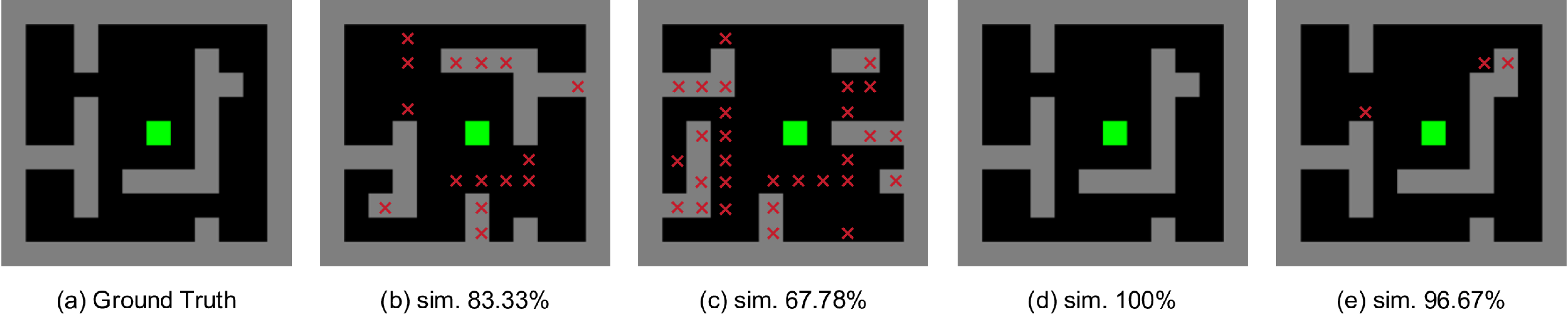}
  \end{minipage}
  \vspace{-0.4cm}
  \caption{(a) The ground truth map. (b, c) Two search results recovered from deterministic agent (DQN). (d, e) Two search results recovered from stochastic agent (Policy Gradient, PG). Cells mismatched with the ground truth are marked with red crosses.}
  \label{garesultdemo}
\end{figure*}

\begin{figure}[!tbp]
  \centering
  \begin{minipage}[b]{0.47\textwidth}
  \includegraphics[width=\textwidth]{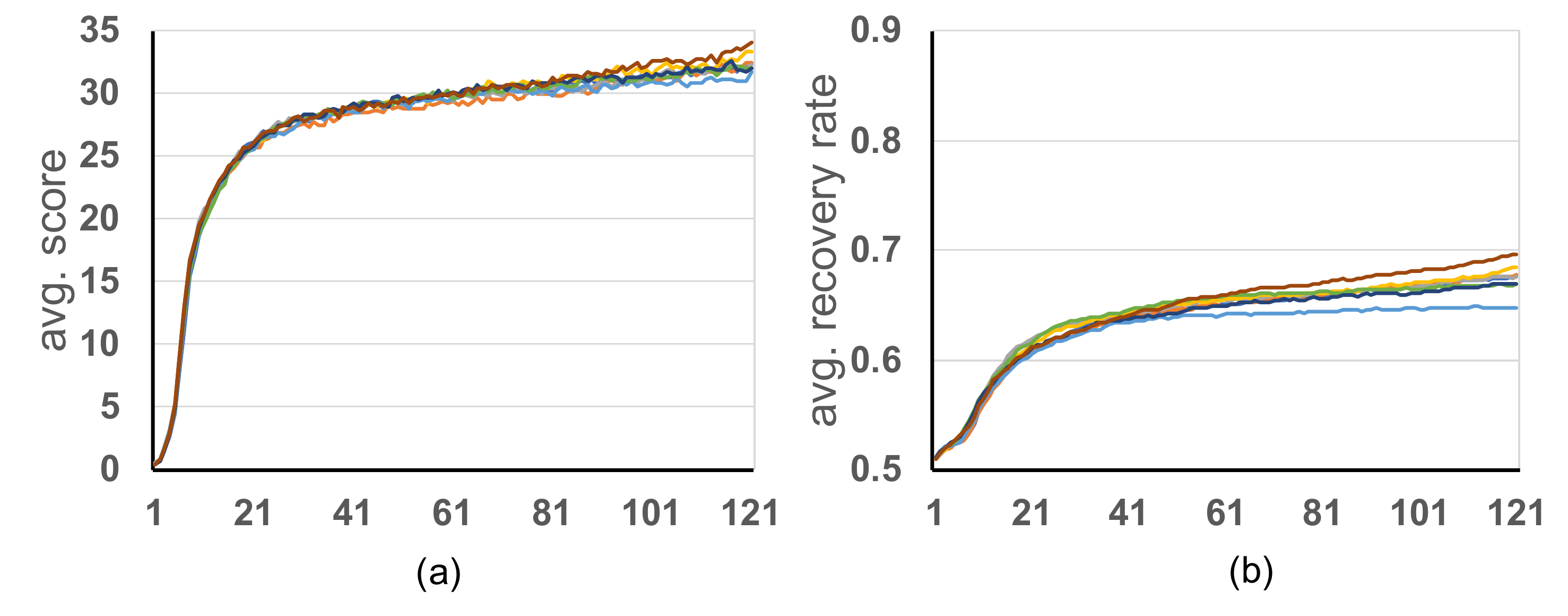}
  \end{minipage}
  \caption{Curves of fitness score in the 8 GA search runs with
  DQN agents. Different colors indicate different seeds.
  The x-axis indicates the number of GA iterations. (a) The curve of population average score. (b) The curve of population average recovery rate (accuracy of free space or obstacle
  prediction).}
  \label{gadcurve}
  \vspace{-0.5cm}
\end{figure}

\begin{figure}[!tbp]
  \centering
  \begin{minipage}[b]{0.47\textwidth}
  \includegraphics[width=\textwidth]{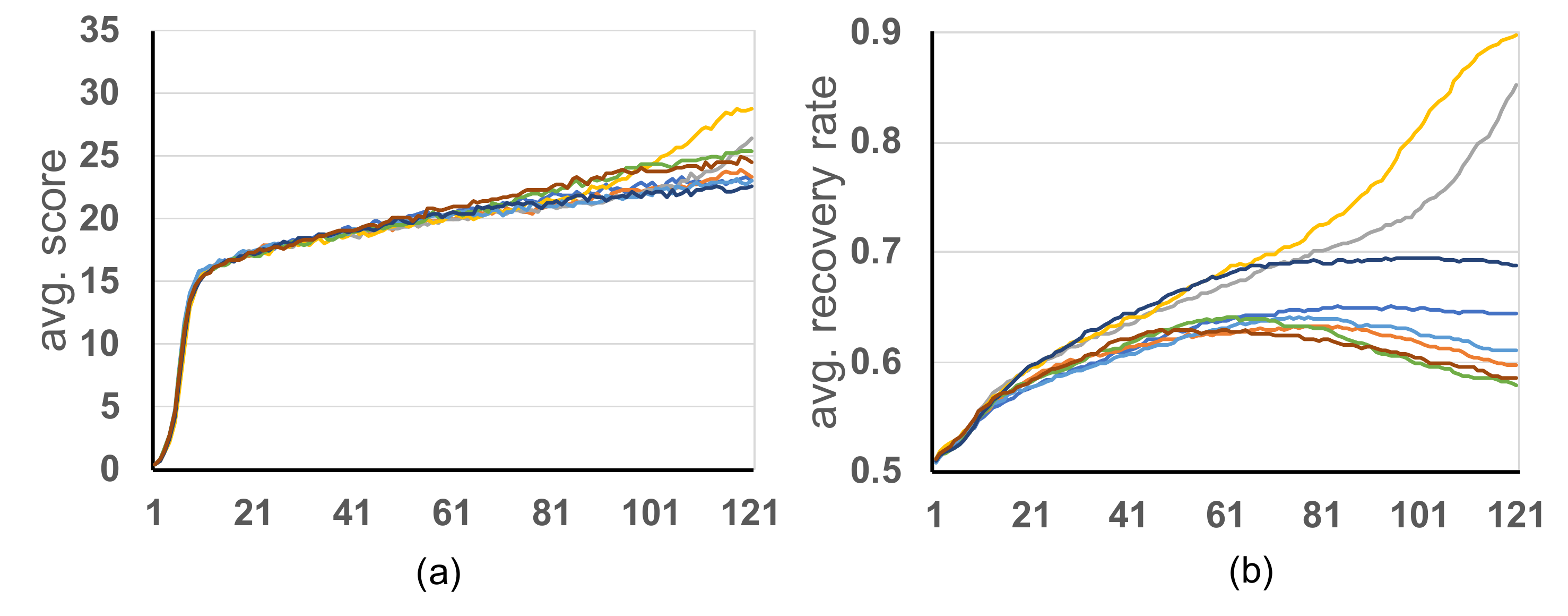}
  \end{minipage}
  \caption{Curves of fitness score in the 8 GA search runs with Policy Gradient (PG) agents. Different colors indicate different seeds. The x-axis indicates the number of GA iterations. (a) The curve of population average score. (b) The curve of population average recovery rate.}
  \label{gascurve}
\end{figure}

We consider the case that the agent takes LiDAR as input instead of aerial-view visual data. More specifically, the observation space is composed of distances in positive real value in 8 directions (4 cardinal directions and 4 intermediate directions). The action space contains five actions: move left, move right, move up, move down and stay. 
The reward function is defined as 1 if reaching
the goal, and -0.1 if the agent stays in place or collides with obstacles, and 0 otherwise. The goal location is fixed for each grid map.
RL agents are trained until full convergence on the environment before testing. The detailed map constraints and policy training details are included below.

\textbf{Floor Plan Constraints}. 
All grid maps are designed following some common sense
in real world architecture design. In particular, an $H \times W$ floor plan, together with its boundary walls, should satisfy the following constraints: First, free space (including the goal) grids form a connected graph, which means the smallest navigation
    distance between any two free space grids is finite; Second, there should be one and only one goal position on the grid map, and the goal grid is considered as free space, not
    an obstacle; Third, the thickness of walls must be equal to 1. In other words, there should not be any obstacles of shape $2 \times 2$, or any obstacles containing this shape.
Note that the boundary walls are not considered as part of the map, and are considered as a known prior. 
An example floor plan is shown in Figure~\ref{gridmapdemo}.


\textbf{Policy Training Details}. The target policies under
a set of specific floor plan designs are trained using 
DQN~\cite{mnih2015human}. The policy network structure is shown in
Table~\ref{archlidar}. We train the agents using
the aforementioned reward design.
For using DQN, we train
the agents using an epsilon-greedy exploration schedule
with exploration rate decreasing from 1 to 0.02 from
step 0 to step 100,000. We also train agents
with vanilla policy gradient methods to get
stochastic policies. The policy gradient agents
share the same network architecture, the same
environment reward functions, and the same
exploration schedule with DQN.

\begin{table}[h]
	\caption{Architecture of network used for LiDAR input agents. ReLU activation are applied after each linear layer except the last one.}
	\label{archlidar}
	\centering
	\begin{tabular}{ccc}
		\toprule
		Type & Input Dim. & Output Dim. \\
		\midrule
		Linear &  8& 64 \\
		Linear & 64& 64 \\
		Linear & 64&  5 \\ 
		\bottomrule
	\end{tabular}
\end{table}

\subsection{Attack Implementation}

Under this setting, we implement the GA based search method to recover the map structure. We randomly generate 20 test floor plans of size $7\times 7$
according to the constraints mentioned earlier. For DQN, the fitness
score (similarity score) measures the fitness of a grid map's
transition dynamics $\mathcal{T}$ to a given policy $\pi_{\textrm{target}}$.
The similarity metric $\Delta$ for deterministic policy (DQN) is 0 if the action selections disagree between the target policy and current policy and it's 1 if they agree; for stochastic policies with the policy gradient method (PG), we use the L2 distance between their action probability distribution, and we set a threshold $\epsilon=0.02$ such that if the
action probability distribution's L2 distance
between the two policies is smaller than $\epsilon$, then the similarity metric returns 1, otherwise it returns 0. Therefore maximizing
the score is equivalent to minimizing the policy difference.



\begin{figure}[!tbp]
  \centering
  \begin{minipage}[b]{0.48\textwidth}
    \includegraphics[width=\textwidth]{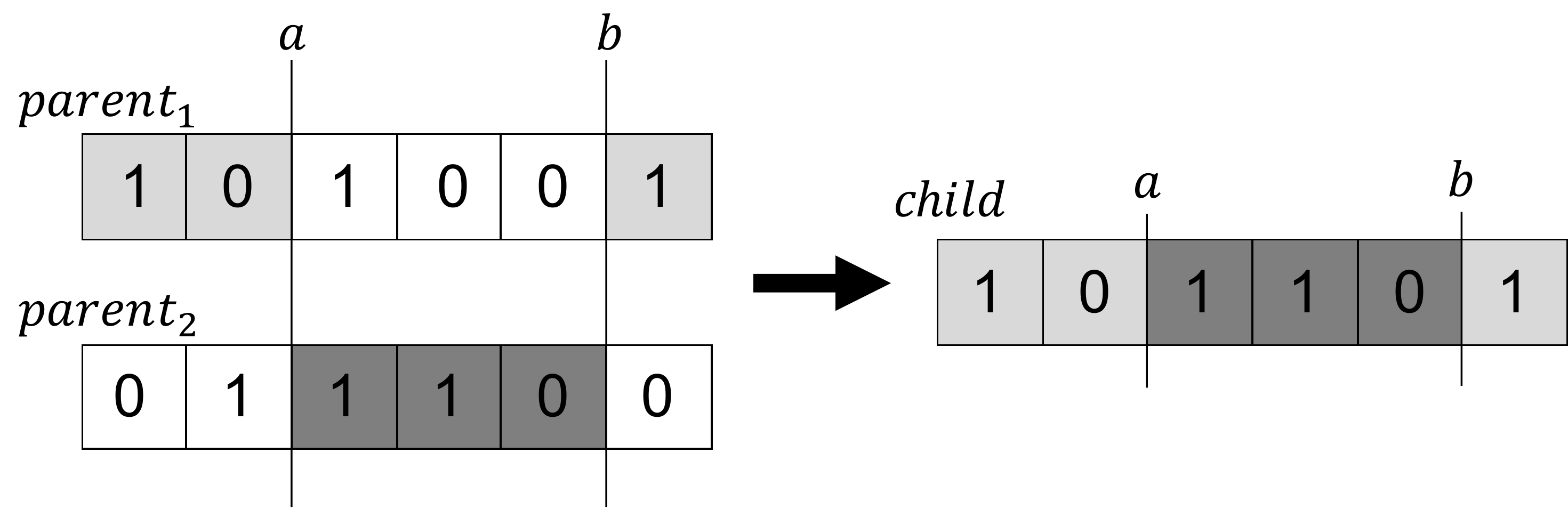}
  \end{minipage}
  \caption{An example of two-point crossover.}\vspace{-0.5cm}
  \label{crossover}
\end{figure}

\textbf{Genetic Algorithm Details}.
In our GA variant, a population $P$ size $L$ is evolved iteratively. The population is composed of candidate map solutions $m_i$, represented by a 0-1 vector (0 for empty and 1 for wall). At every iteration (called generation), each $m_i$ is evaluated using Equation~\ref{eq:eq2}, produces a fitness score $F_i$, and is sorted according to the score. The top $n$ candidates (elite population) are kept to the next generation. The other candidates are each generated by two candidates (called parents) of the last generation. Our GA variant selects parents by randomly selecting two candidates and choosing the one with higher score. Then a two-point crossover of the selected parents is used to generate their child candidate. Random mutation is applied to the resulting child candidate with a mutation rate $\beta$. 
The operators used in our GA variant include: First, crossover operator. We use two-point crossover in our implementation. For the two-point crossover on vectors of length $l$, two random crossover positions $a, b\ (0 \leq a \leq b \leq l)$ are generated for each parent pair. The resulting child candidate is a combination of $parent_1$[0, a), $parent_2$[a, b) $parent_1$[b, END]. This process is shown in Figure~\ref{crossover}. Second, mutation operator. We preset a fixed mutation rate $\beta$ (in our experiments, set to 0.05). For each child vector generated by the crossover process, all bits within their vector have the chance to flip with probability $\beta$.

Based on the fitness scores defined previously, we implement the introduced GA search method. As previous papers have shown that GA tends to converge to a local minimum~\cite{rocha1999preventing}, we run the proposed GA search multiple times with 8 different random seeds. The highest-scored one is selected as our final search result. In addition, we compare GA search with two baseline search methods including random search and RL-based search.

\textbf{Random Search:} The most direct method that can be applied here is random search. In each search iteration, we randomly generate a grid map and calculate its score according to Equation~\ref{eq:eq2}, and consider the map with the highest score as our solution. While there is no guarantee that random search will find a solution, a time limit (5000 s) is set for the search algorithm.

\textbf{RL-based Search:} Reinforcement learning has recently been used to solve combinatorial optimization problems and
search problems~\cite{DBLP:conf/icml/BelloZVL17}. Here we use reinforcement learning as a baseline to search for a Grid World map. The MDP for RL is defined as such:
the state space is all the possible map configurations; the action space is discrete, and consists of overall $H\times W$ 
actions for a map of $H\times W$ shape; the reward function is defined the same as in GA algorithm's fitness
score; the terminal condition is when the number of episodic step exceeds 100 steps. We use DQN as the reinforcement learning search
algorithm and use epsilon-greedy policy for exploration. More specifically, for the testing
maps of $7\times7$ size, the input to the DQN network is a 49 dimension binary vector representing
the current guessed map, and the outputs are the Q-values of 49 different actions, where each action
indicates that specific location to be changed from obstacle to free space or from free space to obstacle. The goal position will not change and it's considered to be known as prior information.
The RL algorithm runs for 250,000 steps before terminating, and the exploration rate decays linearly
from 1 to 0.02 from step 0 to step 240,000. The running time is also set to be 5000 s. We use a three layer fully connected deep neural network to train this RL based search method, with each layer's output size to be 64 except the last layer's output size, which is 49.


Table~\ref{lidarresults} summarizes the accuracy of floor plan recovery by different methods (the best recovery rate). As we can see, with a similar running time, the GA based search method outperforms the other two
methods and achieves high accuracy in terms of map structure recovery. 
Therefore, our attack method is able to recover the map structure and poses a great threat to the privacy of DRL.

To have a better understanding of the GA results, we take the floor plan in Figure~\ref{gridmapdemo} for example. Figure~\ref{garesultdemo} is the visualization of the final search results from the GA algorithm. The curves of population average score and average recovery rate are plotted in Figure~\ref{gadcurve} and Figure~\ref{gascurve}. As can be seen, the quality of the recovered maps from the policy gradient (PG) agent are significantly better than those from the DQN agent. This is intuitive because the former agent gives much more information than the latter one. It is easier to tell if an agent has been trained on some states with the complete probability information. If we compare the curves of two cases (DQN vs PG), we can see the population average score from DQN is higher and more concentrated than from PG, but all runs of DQN agents show a lower average recovery rate (approximately 70\%), while some runs of PG agents could reach recovery rate higher than 90\%. This is
due in part to the score function used for PG. In DQN,
the score function only carries binary information (the
same selected action or different actions), while in PG,
the whole action probability distribution is compared,
which provides more information about the uncertainty
in action selection.

\begin{table*}[!htbp]
\centering
  \caption{Map recovery rate and candidate inference accuracy.}
  \begin{tabular}{c c c c c c}
  \toprule
  Environment & Agent & Task & Method & Recovery Rate & Run Time (s) \\
  \midrule
  Grid World & DQN & Transition Dynamics Search & Random Search      & 61.31\% & 5000  \\ 
  Grid World & DQN & Transition Dynamics Search & RL-based Search    & 81.63\%      & 5000  \\ 
  Grid World & DQN & Transition Dynamics Search & Genetic Algorithm & 89.55\% & 4511  \\ 
  \midrule
  Grid World & PG & Transition Dynamics Search & Random Search      & 68.55\% & 5000  \\ 
  Grid World & PG & Transition Dynamics Search & RL-based Search    & 71.42\% & 5000  \\ 
  Grid World & PG & Transition Dynamics Search & Genetic Algorithm & 95.83\% & 4063  \\ 
  Hopper & PPO & Candidate Inference & SVM & 81.30\% & - \\
  Half Cheetah & PPO & Candidate Inference & SVM & 82.30\% & - \\
   Bipedal Walker & PPO & Candidate Inference & SVM & 88.90\% & - \\
   TORCS & DQN & Candidate Inference & SVM & 94.30\% & - \\
  \bottomrule
  \end{tabular}
  \label{lidarresults}
\end{table*}

\section{Which Bot Did You Use?}

In this section, we present results showing how an attacker could inversely acquire private information using candidate inference with shadow policies under the second scenario we defined previously. The hypothetical attack scenario happens when an attacker tries to identify which version of the robot was used based on the released policy. 
We conduct the experiments on a suite of continuous control benchmarks including the RoboSchool version of MuJoCo tasks Hopper and Half-Cheetah, Box2D environment bipedal walker, and TORCS, a car racing simulator. 

\textbf{Hopper.} The hopper is a monopod robot with four links corresponding to the torso, thigh, shin, and foot and three actuated joints. The goal for the robot is to learn to walk on a plane without falling over by applying continuous-valued forces to its joints. The reward at each time step is a combination of the progress made and the costs of the movements, e.g. electricity, and penalties for collisions. Three environment parameters can be varied: 1. torso density 2. sliding friction of the joints 3. power of the forces applied to each joint. We construct 6 candidate configuration by alternating these parameters. Namely, they are `LightTorso', `HeavyTorso', `SlipperyJoints',`RoughJoints',`Weak' and `Strong'. 

\textbf{Half Cheetah.} Half-cheetah is a bipedal robot with 
eight links and six actuated joints corresponding to the 
thighs, shins, and feet. The goal, reward structure, 
parameters and the way we construct candidate configurations are the same as those in Hopper.

\textbf{Bipedal Walker.} Bipedal Walker is a bipedal robot. 
The task is to move to the far end, and the agent
gets a reward if it succeeds.
The state space consists of the hull angle speed, angular velocity, horizontal
speed, vertical speed, position of joints and joint angular 
speeds, legs contact with ground, and 10 LiDAR rangefinder
measurements. We construct 9 candidate configurations by changing the value of the hull angle speed. 

\textbf{TORCS.} TORCS is an open-source car racing simulator
containing multiple categories of environments with different 
road conditions and terrain. We use the Michigan Way 
environment, which is a racing road with sunny weather. 
The simulator comes from the modified TORCS simulator
from \cite{pan2017virtual}.
Reinforcement learning models trained on TORCS get input as
RGB images from the simulator and can execute 9 different
actions, including move forward and keep current speed,
move forward and accelerate, move forward and decelerate,
turn left and keep current speed, turn left and accelerate,
turn left and decelerate, turn right and keep current speed, 
turn right and accelerate, turn right and decelerate. The 
reward function is defined as $s*(\cos{\theta}-|\sin{\theta}|)/40$
when there is no collision, and -2.5 when there are collisions,
where $s$ is the current speed of the car, and $\theta$ 
is the angle between the speed direction and road direction. The game terminates when the number of steps exceeds 1000 or when the the car collides into obstacles 3 times. We construct 5 configurations with different surface friction coefficients of the road, side walk, and fences. We change the friction of the surfaces from very slippery to slippery to normal, then to rough and very rough.
\begin{figure}[h]
  \centering
  \begin{minipage}[b]{0.2\textwidth}
    \includegraphics[width=0.8\textwidth]{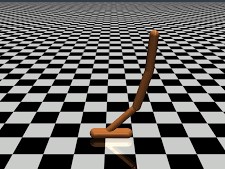}
  \end{minipage}
  \hspace*{1cm}
  \begin{minipage}[b]{0.2\textwidth}
    \includegraphics[width=0.8\textwidth]{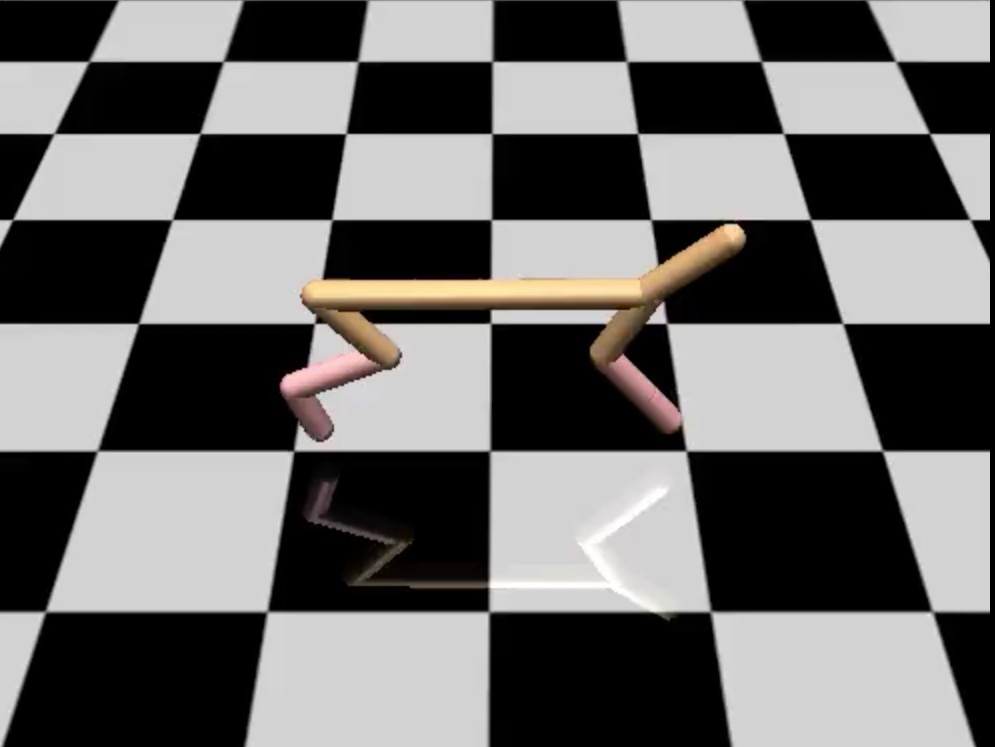}
  \end{minipage}
  \caption{Hopper and Half-Cheetah environments.}
\end{figure}

\begin{figure}[h]
  \centering
  \begin{minipage}[b]{0.2\textwidth}
    \includegraphics[width=0.8\textwidth]{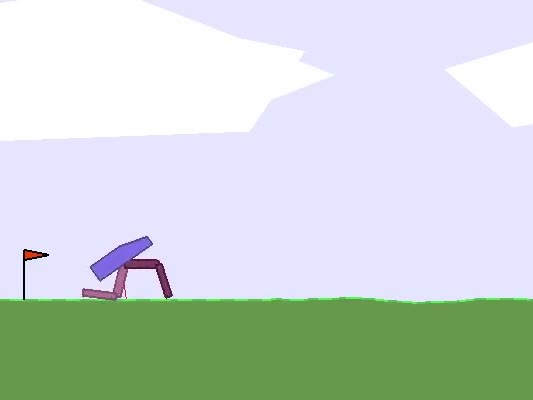}
  \end{minipage}
  \hspace*{1cm}
  \begin{minipage}[b]{0.2\textwidth}
    \includegraphics[width=0.8\textwidth]{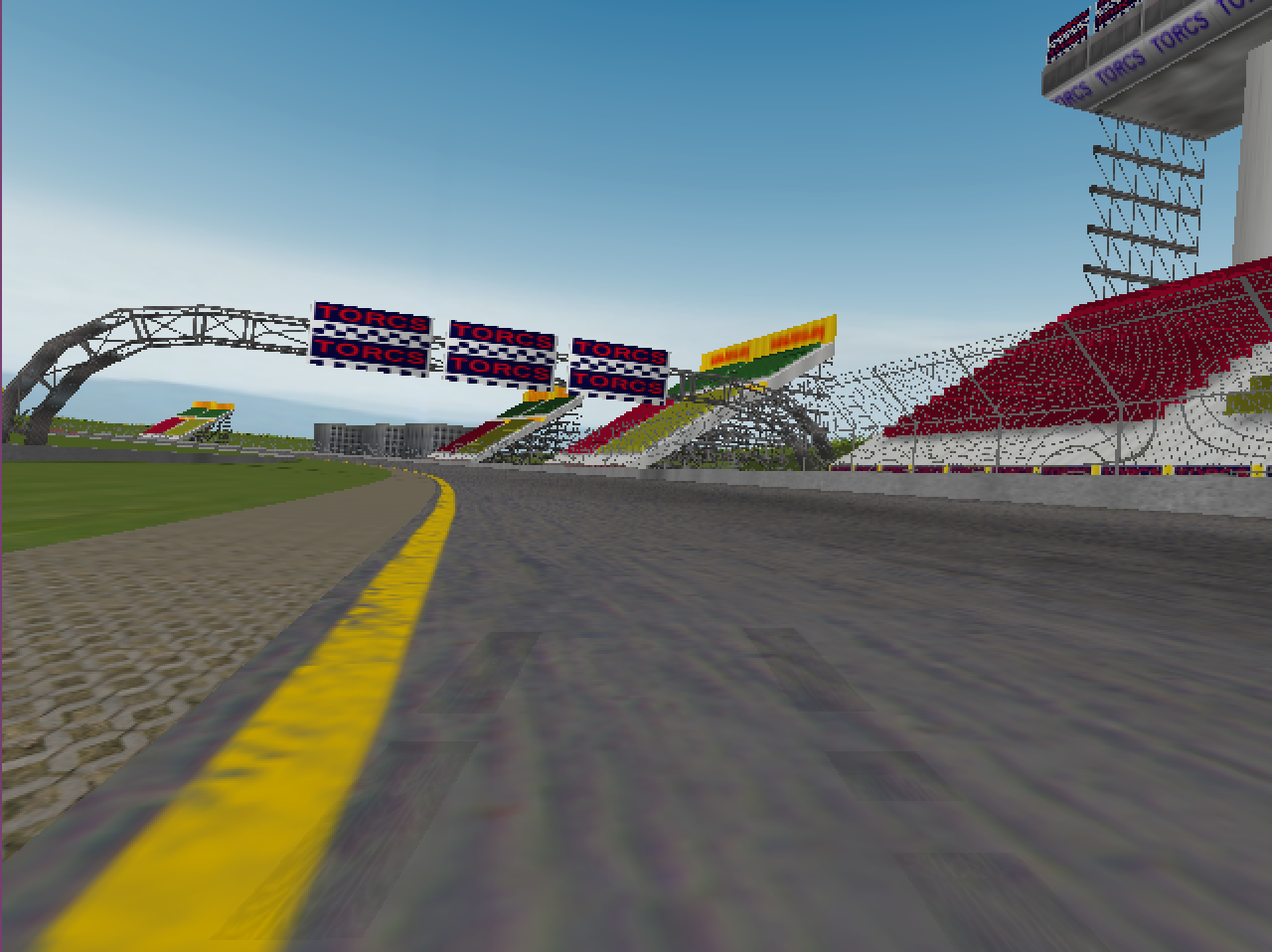}
  \end{minipage}
  \caption{Bipedal Walker and TORCS environments.}
\end{figure}

Note that the parameters involved in constructing candidate environments are important private information. For example, in the TORCS environment, the surface friction coefficients can help to infer the material used to construct the road or indoor environment. Hypothetical attackers
can use such information to potentially know more about the background of the autonomous driving environment. The road's
friction coefficient can also reflect the weather conditions
where the policy is trained. 

For Hopper and Half-Cheetah environments, we use Proximal Policy Optimization (PPO) algorithm \cite{DBLP:journals/corr/SchulmanWDRK17} to train policies.
For each dynamics configuration, we train for 10,000 episodes
until convergence. For bipedal walker, we use PPO to train policy on each configuration for 3,000 episodes until convergence. While training the PPO policy, we use a random environment seed for each episode to avoid overfitting to the seed. The policy and value functions are multi-layer perceptrons (MLPs) with two hidden layers of 64 units each and hyperbolic tangent activations; there is no parameter sharing. 

For the TORCS environment, we use DQN to train the policy to
be attacked. For all configurations, we train for one million steps with exploration rate decaying from 1 to 0.02 from step 0 to step 500,000, and keep the
exploration rate at 0.02 from 500,000 steps on. The DQN network consists of 2 convolutional layers and 3 fully connected layers, and the input image size is $256\times256\times3$. The network architecture is
included in Table~\ref{torcstable}.

\begin{table}[h]
	\caption{Architecture of network used for training TORCS. ReLU activations are applied after each linear or convolutional layer 
	except the last linear layer.}
	\label{torcstable}
	\centering
	\begin{tabular}{cccc}
		\toprule
		Type & Kernel & Stride & Output Channels \\
		\midrule
		Conv. &  8 & 4& 32 \\
		Conv. &  4 & 2& 64 \\
		Conv. &  3 & 1& 64 \\
		\midrule
		Linear & - & - & 512 \\
		Linear & - & - & 9 \\
		\bottomrule
	\end{tabular}
\end{table}

For each configuration, we train policies with 32 random seeds on the training algorithms, splitting the different seeds into training and testing datasets. Out of the policies under
32 seeds, 8 of them were used to generate the training samples to learn the classifiers and the rest were used to do evaluation. The accuracy is computed by averaging accuracy over all configurations. For each task, we have $N$ different candidate dynamics, and $N$ varies across different tasks. For Hopper and Half-Cheetah, we have 6 candidates, and for Bipedal Walker we have 9, for TORCS, we have 5. We use a linear support vector machine (SVM) classifier to predict the label for a given policy.

\begin{figure}
 \centering
 \begin{minipage}[b]{0.5\textwidth}
 \centering
 \includegraphics[width=0.90\textwidth]{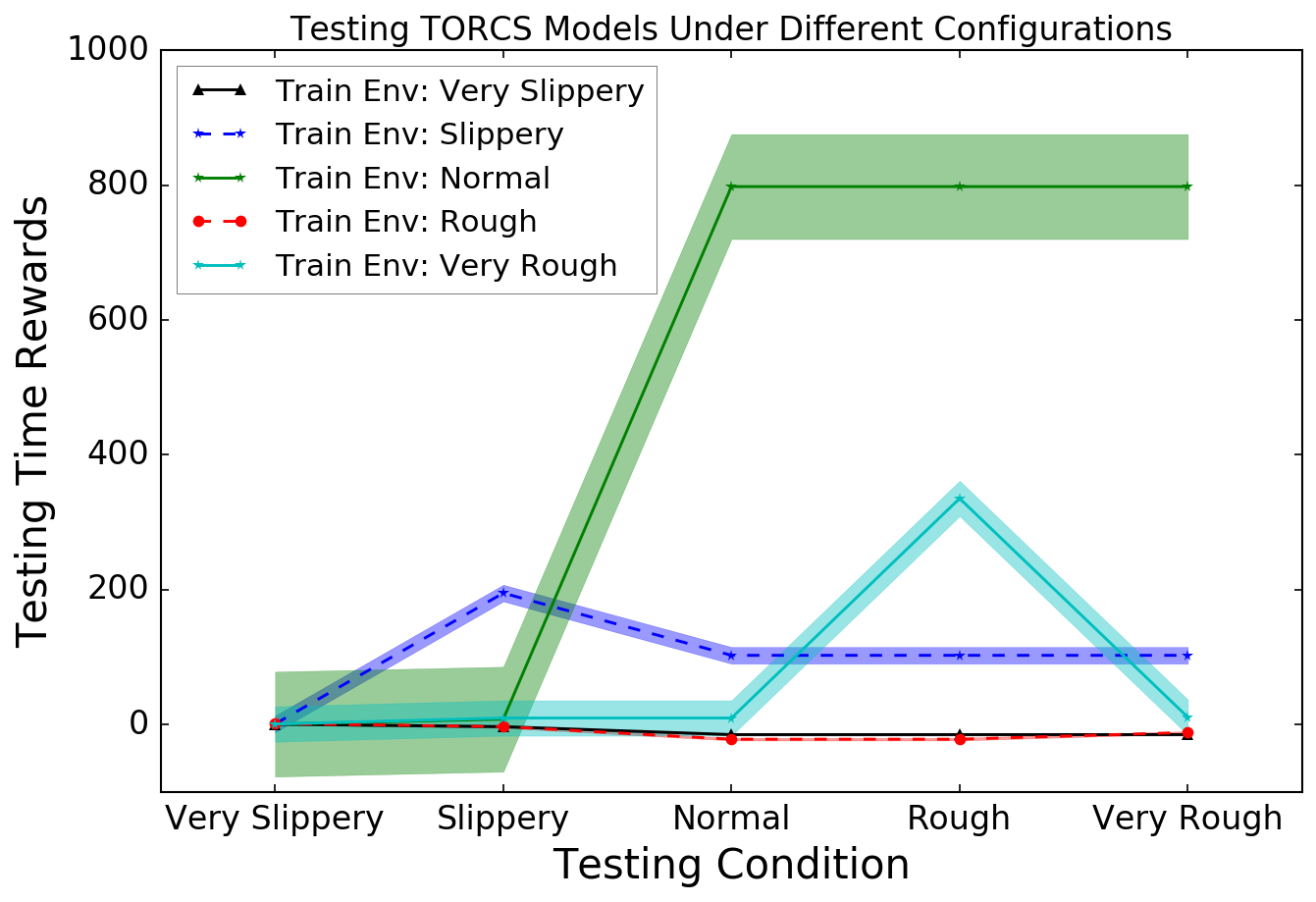}
 \caption{TORCS dynamics candidate testing performance.}
  \label{fig9:torcs}
 \end{minipage}
\end{figure}

In Figure~\ref{fig9:torcs}, we show the test time
performance of models trained in different dynamics in the
TORCS environment. It shows that changing the dynamics does change the model's performance in different testing environments. This makes it possible for a SVM classifier to identify the candidate label for a given policy. Meanwhile, the policy does have some generalization ability. Take the policy trained on normal environment for example: it has the same performance on normal, rough and very rough roads. This indicates that the environment
where the policy performs the best may not be
exactly the environment where it was trained on. On the other hand, Table ~\ref{lidarresults}
shows that a trained SVM classifier can identify the pattern of performances for each policy, and use this to figure out where each policy was trained.

\begin{figure}
 \centering
 \begin{minipage}[b]{0.5\textwidth}
 \centering
 \includegraphics[width=0.9\textwidth]{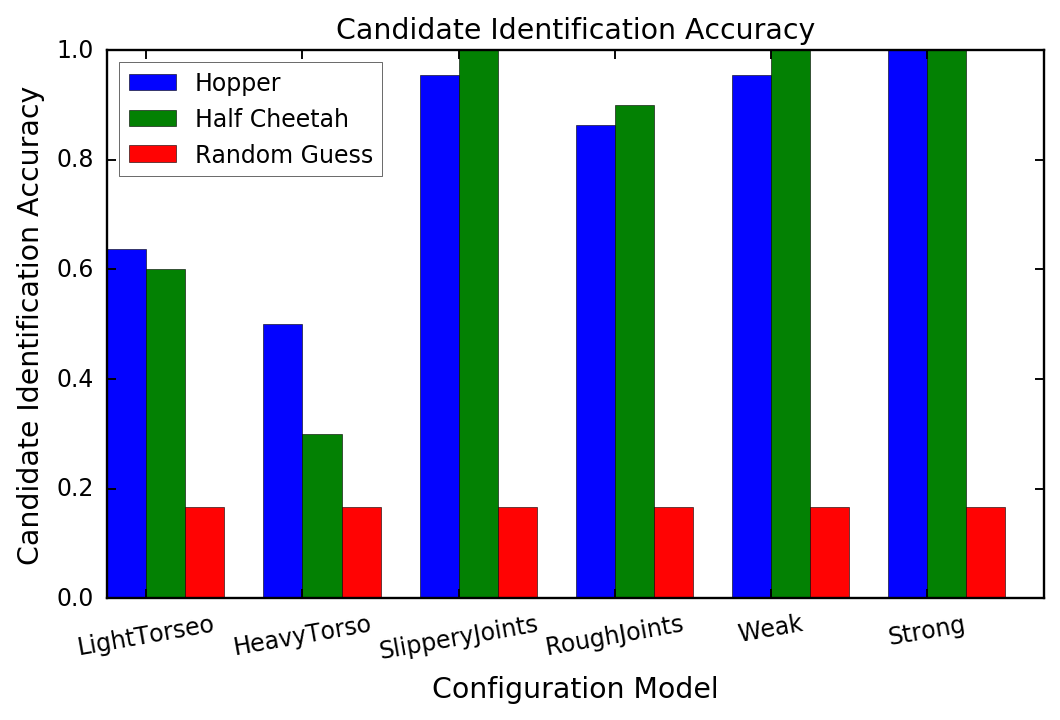}
 \end{minipage}
  \caption{Hopper and Half-Cheetah candidate inference accuracy for different candidates.}
 \label{fig:10}
\end{figure}

In Figure~\ref{fig:10}, we present the accuracy of identifying the underlying training candidate in both the Hopper and Half Cheetah environments. It shows that some candidates are easier to identify than others. Interestingly, this pattern is shared between Hopper and Half Cheetah. 


We report our candidate inference accuracy in table \ref{lidarresults} for all environments. Given the policy rollout
performance in all environments with
different dynamics, we classify this
policy into the environment where it
was originally trained. It illustrates that using the candidate inference with shadow polices method, we achieve high accuracy across all environments. It would be possible for an attacker to inversely infer which candidate configuration was used to train the given policy. This potentially poses a privacy-leaking
challenge to existing reinforcement learning algorithms such as PPO and DQN. 

We discuss designing deep
reinforcement learning policies that are
robust to such privacy-leaking attacks. As
mentioned in the related works section,
robust reinforcement learning usually means
training policies that can generalize to
slightly perturbed environments. The purpose
is to train policies that can be robust
to perturbation in the environment to
ensure safety in control. Our work indicates
that if trained RL policies can generalize
to perturbed environments, the policies
may also be robust to privacy-leaking
attacks. For example, in the Grid World
case, if the policy is not memorizing
the environment dynamics but instead
learns to plan no matter what environment
it is deployed in, then it would be almost
impossible to infer the floor plan structure
using our method. In the candidate inference
method, if the trained policies can generalize
to many different environments, including
significantly perturbed environments,
then it will be much harder to infer
which environment appeared in the training
environment. Therefore, our work
provides some insight into designing robust
RL algorithms that can defend against such
privacy-leaking attacks.

\section{Conclusion}

In this work, we have evaluated several approaches to retrieve private information about the environment from well-trained policies under different settings. We evaluated our proposed algorithms under two settings. In the first setting we apply some constraints on the transition dynamics, and proposed to use a genetic algorithm to recover the original transition map in a Grid World environment. In the second setting, we assume we have access to
several candidate dynamics and propose an algorithm to infer which candidate environment is used to train a given policy. Through extensive experiments we show that deep reinforcement learning is vulnerable to potential
privacy-leaking attacks and specific information about the training environment dynamics can be recovered with high accuracy. These findings provide insights for designing reinforcement learning algorithms
that protect the training environment's privacy. Our work also provides reinforcement learning community with a new perspective at the intersection of
RL and its generalizability, privacy and security. 

\subsection*{Acknowledgement}
This work is partially supported by DARPA grant 00009970. We thank Warren He for providing useful feedback.

\clearpage
\newpage

\bibliographystyle{ACM-Reference-Format}  
\balance  
\bibliography{bib}  


\end{document}